\documentclass[conference]{IEEEtran}

\IEEEoverridecommandlockouts
\IEEEpubid{\makebox[\columnwidth]{978-1-7281-4852-6/19/\$31.00~
\copyright~2019 IEEE \hfill} \hspace{\columnsep}\makebox[\columnwidth]{}} 

\usepackage{caption}
\usepackage{cite}
\usepackage[pdftex]{graphicx}
\usepackage[cmex10]{amsmath}
\usepackage[tight,footnotesize]{subfigure}
\usepackage{epsfig}
\usepackage{amssymb}
\usepackage{multirow}

\begin{document}

\title{FVNet: 3D Front-View Proposal Generation for \\ Real-Time Object Detection from Point Clouds}

\author{\IEEEauthorblockN{Jie Zhou, Xin Tan, Zhiwen Shao$^*$\thanks{$^*$ Corresponding author.} and Lizhuang Ma}
\IEEEauthorblockA{Department of Computer Science and Engineering, Shanghai Jiao Tong University, China\\
\{lord\_liang, tanxin2017, shaozhiwen\}@sjtu.edu.cn, ma-lz@cs.sjtu.edu.cn}
}
\maketitle

\begin{abstract}
3D object detection from raw and sparse point clouds has been far less treated to date, compared with its 2D counterpart. In this paper, we propose a novel framework called FVNet for 3D front-view proposal generation and object detection from point clouds. It consists of two stages: generation of front-view proposals and estimation of 3D bounding box parameters. Instead of generating proposals from camera images or bird's-eye-view maps, we first project point clouds onto a cylindrical surface to generate front-view feature maps which retains rich information. We then introduce a proposal generation network to predict 3D region proposals from the generated maps and further extrude objects of interest from the whole point cloud. Finally, we present another network to extract the point-wise features from the extruded object points and regress the final 3D bounding box parameters in the canonical coordinates. Our framework achieves real-time performance with 12ms per point cloud sample. Extensive experiments on the 3D detection benchmark KITTI show that the proposed architecture outperforms state-of-the-art techniques which take either camera images or point clouds as input, in terms of accuracy and inference time. 
\end{abstract}


\begin{figure*}
\begin{center}
    \includegraphics[width=1\linewidth]{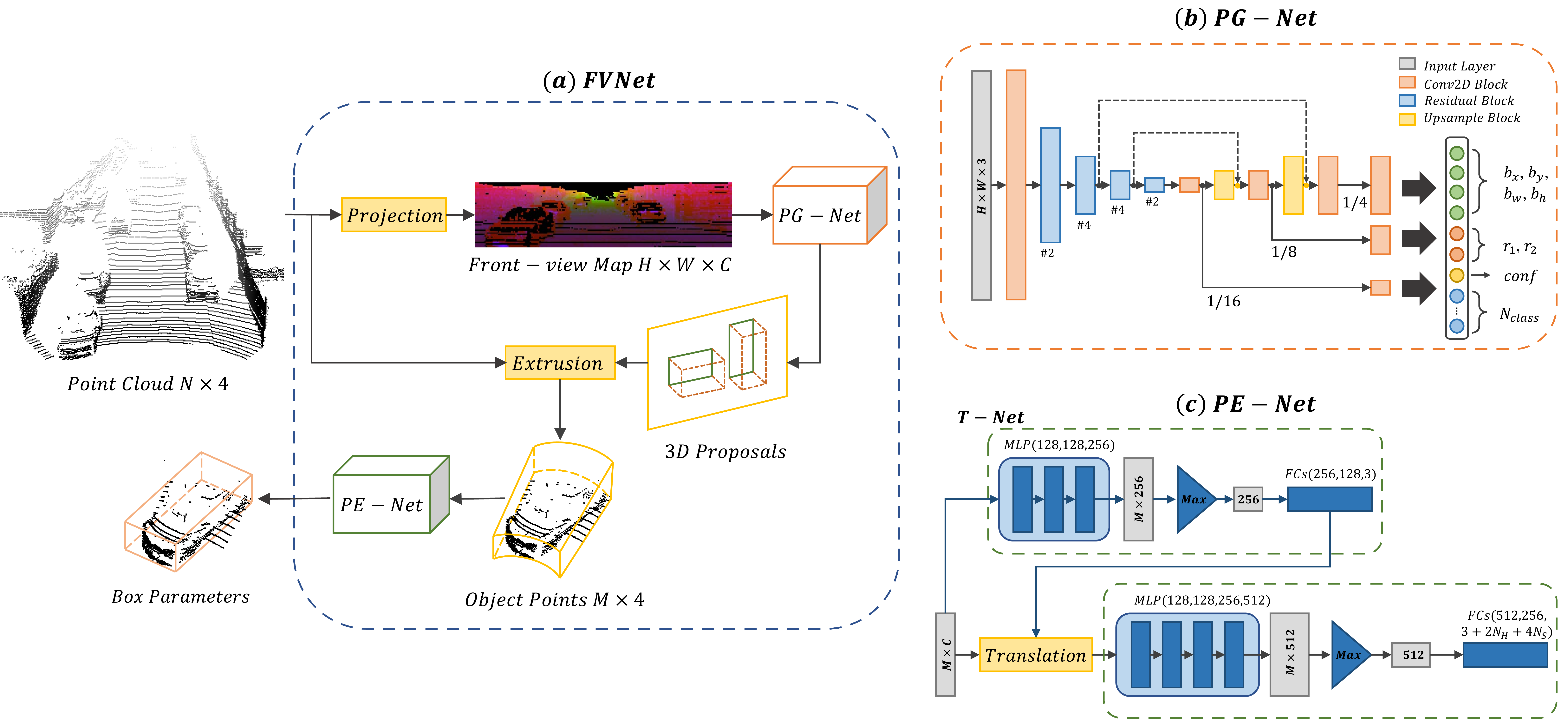}
\end{center}
   \caption{The overview of proposed $(a)$ FVNet. It is composed of two sub-networks: $(b)$ Proposal Generation Network (PG-Net) for generation of 3D region proposals and $(c)$ Parameter Estimation Network (PE-Net) for estimation of 3D bounding box parameters.}
\label{fig:pipeline}
\end{figure*}

\section{Introduction}

3D object detection~\cite{Chen2016Monocular, chen2017multi,ku2018joint, zhou2018voxelnet,shi2018pointrcnn, li2019stereo} has proved to be increasingly important in many fields, such as autonomous driving~\cite{ackerman2016lidar}, mobile robots~\cite{sprunk2013lidar} and virtual/augmented reality~\cite{park2008multiple}. While there have been remarkable progresses in the field of image-based 2D object detection, 3D object detection is far less explored than its 2D counterpart. The purpose of 3D object detection is to understand the geometry of physical objects in 3D space and predict future motion of objects. 
In this paper, we focus on 3D object detection from LiDAR, predicting 3D bounding boxes of objects from raw point clouds.
Compared with camera images, 
LiDAR point clouds are sparse and irregular (highly variable point density) because of the nonuniform sampling of the space, 
the radar coverage, the occlusion and relative attitude. How to encode the point cloud for 3D object detection remains an interesting and open problem.

Deep neural networks usually take vectorized data as input, and the way of vectorizing point clouds is vital to the resulting performance. Some methods~\cite{wang2015voting, engelcke2017vote3deep, zhou2018voxelnet} voxelize point clouds into volumetric grids and then design voxel-wise feature extractors to handle them. However, since the captured point clouds in outdoor scenes are sparse in essence, the voxel grids are very sparse and therefore a large proportion of computation/memory is redundant and unnecessary. To obtain a more compact representation, some methods transform point clouds into 2D maps by projecting point clouds onto the ground plane (Bird's Eye View)~\cite{beltran2018birdnet, zeng2018rt3d, simon2018complex} or depth maps (Front View)~\cite{li2016vehicle, minemura2018lmnet}, and then apply 2D convolutional neural networks on the transformed maps to execute the detection task. The bird's-eye-view representation can preserve the metric space and ease the problem of overlap, as objects usually do not overlap with each other. However, the bird's-eye-view based methods often ignore the size and location along the $Z$ axis. The assumption that objects of interest are constrained to the same ground plane is often unfeasible in practice. Besides, these methods are unfriendly to small objects such as pedestrians and cyclists, for example, BirdNet~\cite{beltran2018birdnet} and Complex-YOLO~\cite{simon2018complex}. 
In comparison to the bird's-eye-view representation, the front-view representation retains more information, which is similar to camera images. Inevitably, front-view representation has the same problem as camera images: the sizes of objects are closely related 
to the distances to scanners, and objects may overlap with each other. It is even more difficult to detect objects on the front-view maps than camera images, due to the loss of texture and color information. Consequently, previous front-view based methods~\cite{qi2018frustum, shin2018roarnet} still utilize camera images to generate 2D proposals. However, these detectors based on camera images often miss objects due to dark lighting or dazzle light, and require high-quality synchronization between 2D camera sensors and 3D LiDAR sensors.

Motivated by the above problems, we propose a novel framework called FVNet for 3D object detection from raw point clouds (see Fig.~\ref{fig:pipeline}). It consists of two sub-networks: proposal generation network to generate 3D proposals and parameter estimation network to estimate bounding box parameters.  With regard to the small object detection, our detector is multi-scale so that it copes with the problem of large variance in object size. 
For the problem of overlap, our detector predicts accurate truncated radial distances to cut the view frustum for each single object. With predicting the truncated distances, we do not need an extra instance segmentation network to determine which points belong to the object of interest, avoiding suffering from computational complexity. To mitigate the problem of loss of texture and color information, we directly operate on the extruded object points to extract the point-wise features for estimating the final bounding box parameters. \textit{Our method depends on point clouds only, without the need of camera images}.
The major contributions of this paper are summarized below.
\begin{itemize}
    \item We propose a novel multi-scale 3D object detector for predicting 3D bounding boxes of objects from raw point clouds.
    \item  We introduce a proposal generation network to generate 3D proposals from front-view maps, without the need of camera images. We extend 2D bounding boxes to 3D bounding boxes by truncating the frustum with truncated radial distances.
    \item Extensive experiments are conducted and the results show that proposed method outperforms state-of-the-art techniques which take either point clouds or camera images as input, in terms of accuracy and inference time. Our framework achieves real-time performance with 12ms per sample.
\end{itemize}


\section{Related Work}

We review the achievements related to this research in retrospect. We first review the CNN-based object detection. We then investigate 3D object detection from point clouds, where the methods are categorized according to the encoding means for point clouds.

\subsection{CNN-Based Object Detection}

Real time and accuracy are key elements for autonomous driving which necessitates high prediction accuracy and efficiency of models. 
2D detectors generally fall into two categories: single-stage detectors and two-stage ones. Two-stage detectors first inference proposals and then refine them, while single-stage detectors predict the final detection results directly. R-CNN~\cite{girshick2014rich} first generated category-independent region proposals and then extracted a feature vector of fixed length from each region with convolutional neural networks. Faster-RCNN~\cite{ren2015faster} further improved the region proposal network and shared the feature representation with the detection network, leading to further gain in both accuracy and speed. Despite that, it is still challenging by applying the typical two-stage pipelines to real-time applications. YOLO~\cite{redmon2016you, redmon2017yolo9000, redmon2018yolov3} and SSD~\cite{liu2016ssd} are two outstanding works with real-time speed. YOLO divided the image into sparse grids and made multi-class and multi-scale predictions per grid cell. SSD additionally used pre-defined anchors to handle large variance in object size and shape. Our method belongs to the single-stage detectors. 

\subsection{3D Object Detection based on Point Clouds}

\noindent\textbf{Projection-Based Methods.}
MV3D~\cite{chen2017multi} hierarchically fused the CNN features extracted from the front view, bird's eye view and camera view to jointly predict object class and regress the oriented 3D bounding boxes. PIXOR~\cite{yang2018pixor} took bird's-eye-view representation as input alone and designed a proposal-free, single-stage detector to output pixel-wise predictions. It, however, assumes that all objects lie on the same ground and cannot handle indoor scenes where multiple objects often stack together in vertical space. BirdNet~\cite{beltran2018birdnet} and RT3D~\cite{zeng2018rt3d} generated region proposals from bird's-eye-view but achieve weak results. VeloFCN~\cite{li2016vehicle} is the first work to project point clouds onto a cylindrical surface. LMNet~\cite{minemura2018lmnet} took the front-view representation as input alone but got unsatisfactory results even on car detection, because of the loss of details. To ease the influence of the lack of details, we directly extract point-wise features from raw point clouds in our stage-2 sub-network (Sec.~\ref{sec:pen}), rather than extracting pixel-wise features from the bird's-eye-view or front-view maps as before.

\noindent\textbf{Voxel-Based Methods.}
Voxel-based methods utilize a voxel grid representation for point clouds and involve differing ways to extract features. Vote3D~\cite{wang2015voting} adopted sliding windows on sparse volumes to extract hand-crafted geometric features for each volume. Vote3Deep~\cite{engelcke2017vote3deep} introduced 3D convolutional neural networks to extract features for each volume. VoxelNet~\cite{zhou2018voxelnet} built three voxel feature encoding (VFE) layers to extract 3D features for the region proposal network. However, the main issues of voxel representation are computational efficiency and memory consumption.

\noindent\textbf{Point-Based Methods.}
F-Pointnet~\cite{qi2018frustum} and F-ConvNet~\cite{wang2019frustumconvnet} directly operated on original point clouds after popping up RGB-D scans, without any loss of data, leading to precise detection. Their generation of object proposals depended largely on camera images. To get more accurate locations of objects, F-Pointnet adopted instance segmentation to classify the points in the view frustum while F-ConvNet divided the view frustum into a sequence of frustums to extract frustum-level features. Our approach also directly operates on raw point clouds, and our proposals are generated from point clouds only, without the need of camera images. We adopt the truncated distances to cut the view frustum for more accurate locations, without the need of point cloud segmentation. 


\begin{figure}
\begin{center}
	\includegraphics[width=1\linewidth]{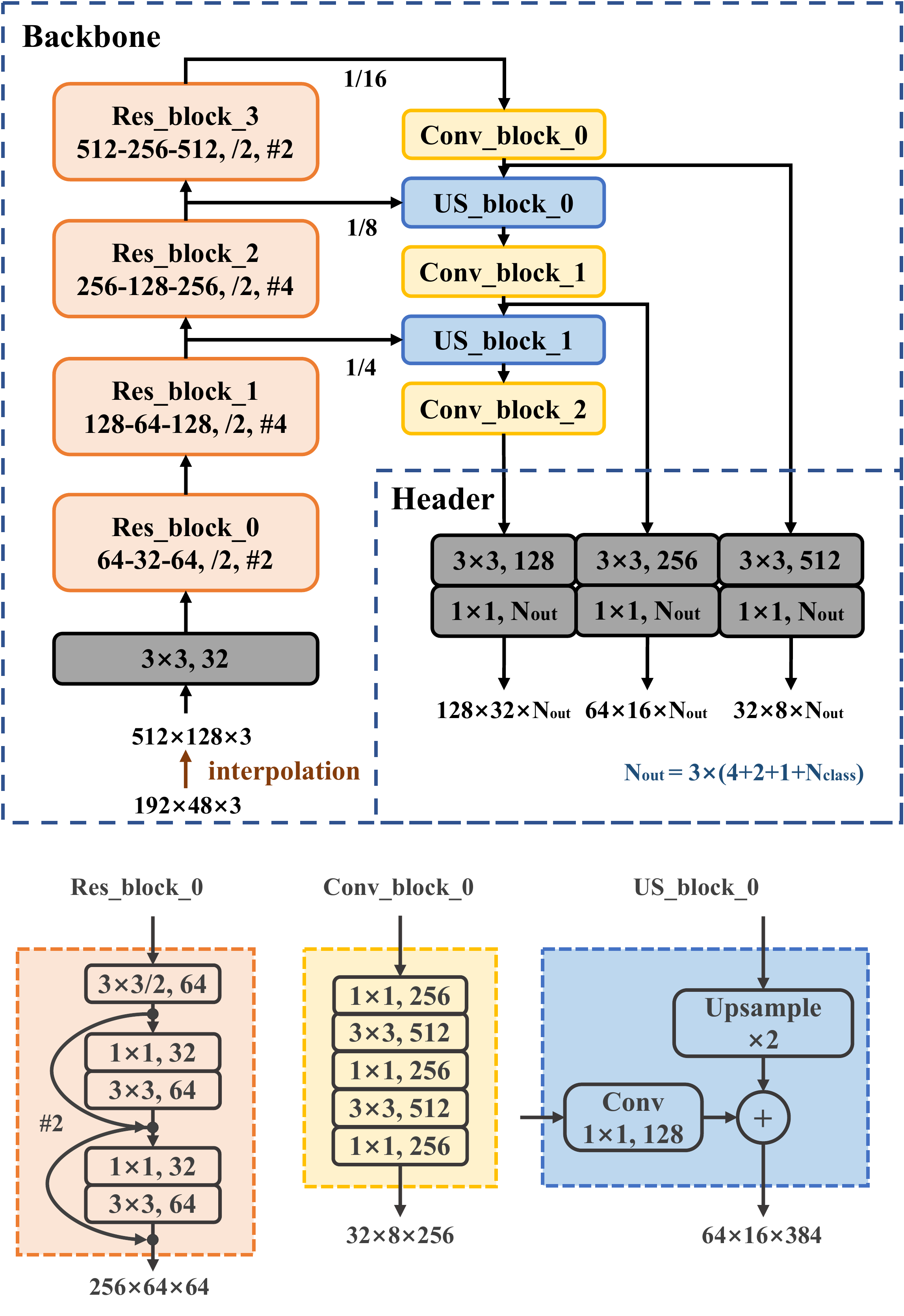}
\end{center}  
   \caption{The architecture of our PG-Net. The bottom shows the details of the residual block, the convolutional block and the upsampling block, respectively.}
\label{fig:dnet}
\end{figure}

\section{Proposed FVNet Framework}

As illustrated in Fig.~\ref{fig:pipeline}, proposed FVNet framework consists of two sub-networks: Proposal Generation Network (PG-Net) and Parameter Estimation Network (PE-Net), respectively illustrated in $(b)$ and $(c)$. Next, these two sub-networks will be introduced in detail.

\subsection{Proposal Generation Network}
\label{sec:pgn}
We design a multi-scale, fully-convolutional neural network for 3D proposal generation, 
and the network architecture is illustrated in Fig.~\ref{fig:dnet}. The backbone network is to extract a general representation of the input in the form of convolutional feature maps. The header network is responsible for task-specific predictions. 

\noindent\textbf{Data Preparation.}
Different from the regular image pixels, point clouds usually distribute sparsely and irregularly, which makes it challenging for direct learning on point clouds. To circumvent this issue, we project raw point clouds onto a cylindrical surface, obtaining a dense representation based on grid. This kind of representation combines the advantages of both image-based and 3D-based methods: (1) the easy extension of 2D convolutional neural networks from camera images to the front-view feature maps, (2) the ability of capturing dependencies across different views because of the depth information stored in the front-view maps. Assume that $p=(x,y,z)^T$ denotes to a certain point in 3D space and $(r, c)$ denotes the 2D map position of its projection. Then, the projection functions are defined as follows:
\begin{equation}
\begin{aligned}
& r = \lfloor\frac{\theta}{\Delta\theta}\rfloor, \quad \theta = \arcsin\frac{z}{\sqrt{x^2 + y^2 + z^2}}, \\
& c = \lfloor\frac{\phi}{\Delta\phi}\rfloor, \quad \phi = \arcsin\frac{y}{\sqrt{x^2 + y^2}},
\end{aligned}
\label{eq: projection}
\end{equation}
where $\theta$ and $\phi$ refer to the \emph{azimuth} angle and the \emph{zenith} angle and $\Delta\theta$ and $\Delta\phi$ are angle units, similar to the definitions from ~\cite{wu2018squeezeseg}.

By projecting a point cloud onto a cylindrical surface with Eq.~\ref{eq: projection}, we can obtain a 3D tensor of size $H\times W\times C$ where $H$ and $W$ are the height and width of the projection map and $C$ is the feature channel number for each pixel. We fill the element at $(r, c)$ with 3-channel features consisting of the height $h=z$, the radial distance $r=\sqrt{x^2+y^2}$ and the LiDAR intensity $i$. We visualize the three channels in RGB (Fig.~\ref{fig:pipeline}$(a)$). Compared to bird's-eye-view projection, projecting point clouds onto a cylindrical surface is based on angle instead of length, getting rid of the need of the scene range restriction.

The bounding boxes of small targets including pedestrians and cyclists may be less than one pixel in size, due to the few details and small size of these objects in the front-view maps. Regarding this, we scale up the original front-view maps to a greater size by enhancing the details with the nearest neighbor interpolation. The nearest neighbor interpolation simply selects the value of the nearest pixel, yielding a piecewise-constant interpolant. Specifically, we enlarge the original front-view maps from $48\times 192$ to $128\times 512$ in our experiments.

\begin{figure}
\begin{center}
\includegraphics[width=1\linewidth]{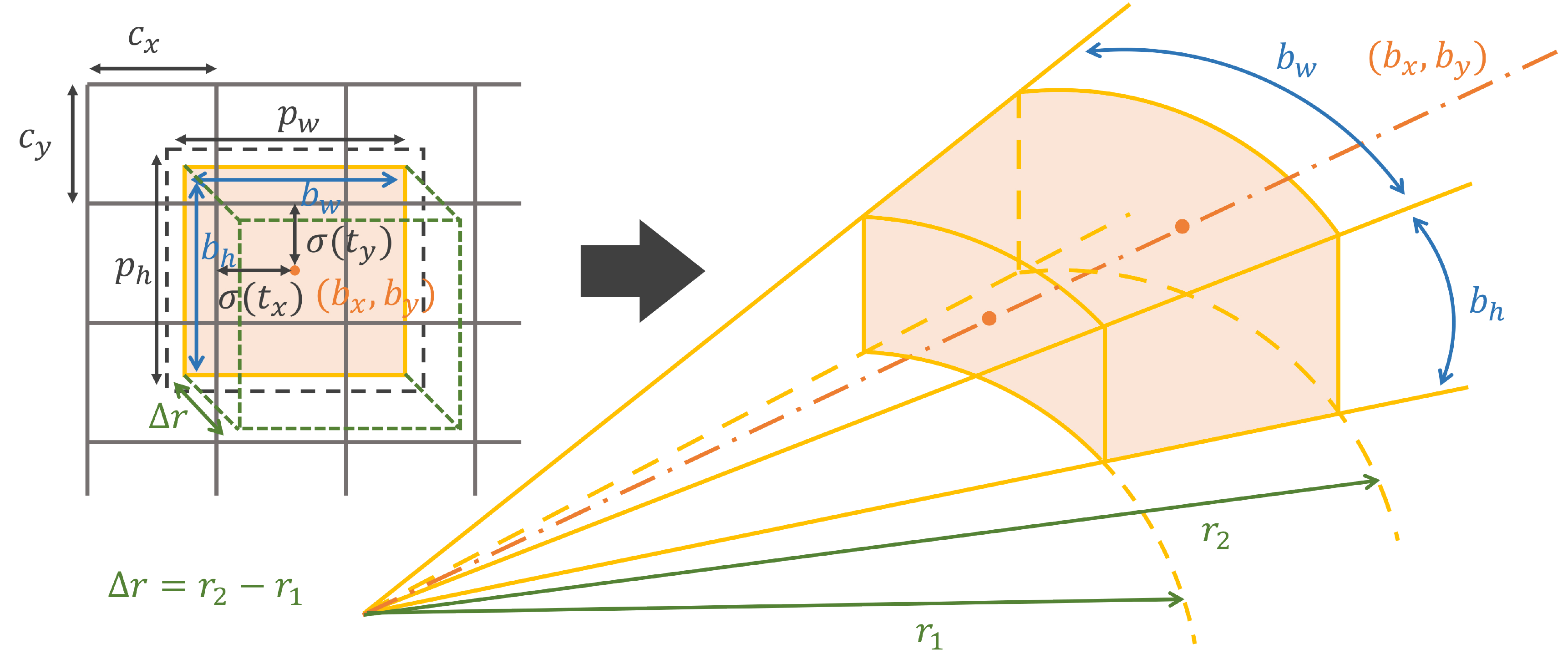}
\end{center}
   \caption{A 3D bounding box and its corresponding cylinder fragment. Left: the 3D bounding box with dimension prior ($P_w$, $P_h$), location prediction ($b_x$, $b_y$) and truncated distances prediction ($r_1$, $r_2$). Right: the corresponding cylinder fragment in 3D space, which is generated by truncating the frustum with two radial distances $r_1$ and $r_2$.}
\label{fig:bbox}
\end{figure}

\noindent\textbf{3D Proposals with Truncated Distances.} 
Thanks to the regularity of front-view feature maps, we can generate our 3D object region proposals on the top of mature 2D object detectors. 2D region proposals usually involve four parameters: the center coordinate $(b_x, b_y)$, the width $b_w$ and the height $b_h$ of the bounding box. Different from 2D region proposals, 3D region proposals need more information to extrude the objects of interest from the frustum, due to the increase of dimension. F-Pointnet~\cite{qi2018frustum} adopted a 3D instance segmentation method (PointNet) to extract the object of interest from a point cloud in the view frustum by classifying the relevant and non-relevant points. Different from their network, \textit{our PG-Net gets rid of instance segmentation}. Our 3D proposals parse the 3D bounding boxes with six parameters $(b_x, b_y, b_w, b_h, r_1, r_2)$. As shown in Fig.~\ref{fig:bbox}, we extend 2D bounding boxes to 3D bounding boxes by truncating the frustum with $r_1$ and $r_2$. $r_1$ and $r_2$ denote the front and back truncated radial distances, respectively. Besides, we adopt a confidence score $conf$ and $N_{class}$ class scores for location and classification.

Following YOLOv3~\cite{redmon2018yolov3}, we use dimension clusters to generate pre-defined bounding box priors. We cluster the width and height of bounding boxes with $K$-means algorithm to achieve $K$ reasonable bounding box priors. Our PG-Net predicts $K/N_{scale}$ bounding boxes at each cell in each output layer, where $N_{scale}$ is the number of output layers. We set $N_{scale}$ to $3$ in our experiments to favor different object sizes (large, medium and small). The network predicts $4+2+1+N_{class}$ values: $t_x$, $t_y$, $t_w$, $t_h$, $t_{r_1}$, $t_{r_2}$, the confidence score $conf$ and $N_{class}$ class scores for each bounding box. We predict the center coordinates relative to the grid coordinates. Suppose each grid cell has an offset $(c_x, c_y)$ from the top left corner and a bounding box prior is of width $p_w$ and height $p_h$, the predictions can be computed as
\begin{equation}
\begin{aligned}
& b_x = \sigma(t_x) + c_x, \ \quad b_y = \sigma(t_y) + c_y,\\
& b_w = p_w e^{t_w}, \quad\qquad b_h = p_h e^{t_h},\\
& r_1 = t_{r_1}R, \  \ \ \quad\qquad r_2 = t_{r_2}R,
\end{aligned}
\label{eq: box}
\end{equation}
where $R$ represents the maximum distance that LiDAR sensors can reach. $\sigma(\cdot)$ refers to the sigmoid function for constraining the coordinate predictions to fall in $[0, 1]$.

We predict a confidence for each bounding box. When a bounding box prior has more overlap with a ground-truth bounding box than other priors, its confidence is $1$. When a bounding box prior does not overlap most but its overlap does exceed a certain threshold (e.g., $0.5$), we ignore the prediction and its confidence is $0$. As a result, only one bounding box prior is allotted to a ground-truth bounding box. 

\noindent\textbf{Backbone Network.} 
With a concept similar to feature pyramid networks~\cite{lin2017feature}, our backbone network extracts features from three different scales for objects of different sizes. Our backbone network architecture is shown in Fig.~\ref{fig:dnet}. 
It has a downsampling factor of $16$ for the purpose of fewer layers in high resolution and more layers in low resolution. The backbone network composes of $4$ residual blocks~\cite{he2016identity}, $2$ upsampling blocks and some convolutional layers. The first convolutional layer of each residual block possesses a stride of $2$ for downsampling the feature map.  The numbers of skip connections are $2$, $4$, $4$ and $2$, respectively. Upsampling blocks are employed to get three final feature maps, with $4\times$, $8\times$ and $16\times$ downsampling, respectively.

\noindent\textbf{Header Network.}
The header network takes the three final feature maps from the backbone network as input, and outputs a 3D tensor for each branch. In our implementation, we predict three bounding boxes at each scale. Thus, the output tensor is $M\times N \times[3 \times (4+2+1+N_{class})]$ for $4$ bounding box offsets, $2$ truncated distances, $1$ confidence prediction and $N_{class}$ class predictions. 

\noindent\textbf{Stage-1 Loss Function.}
The multi-task loss $L_{stage-1}$ for our PG-Net extends from the common loss for 2D bounding boxes. We apply the binary cross entropy loss on the location output $p_{coord}$ (the center coordinates $b_x$ and $b_y$), the confidence output $p_{conf}$ and the classification output $p_{cls}$.  We apply a Huber loss on the regression output $q$ for the width $b_w$, length $b_h$ and truncated distances $r_1$ and $r_2$.
\begin{equation}
\begin{aligned}
L_{stage-1} = &\lambda_{coord}B(p_{coord}, y_{coord}) +\\
&\lambda_{conf}B(p_{conf}, y_{conf})+  \\
&\lambda_{cls}B(p_{cls}, y_{cls}) + \lambda_{reg}H(q-y_{reg}) ,
\end{aligned}
\end{equation}
where $B(p,y)$ refers to the binary cross entropy loss function and $H(x)$ indicates Huber loss function. $\lambda_{coord}$, $\lambda_{conf}$, $\lambda_{cls}$ and $\lambda_{reg}$ are weight coefficients, and $y_{(\cdot)}$ denote the corresponding ground-truths. Considering that cars and persons may overlap in the front-view maps, we use multi-label binary cross entropy loss.  These two loss functions can be computed as
\begin{equation}
\begin{aligned}
B(p,y)=&-\frac{1}{n}\sum_i(y[i] \log(p[i]) +\\
&(1-y[i]) \log(1-p[i])),
\end{aligned}
\end{equation}
\begin{equation}
H(x)=\left\{
\begin{array}{ll}
\frac{1}{2}x^2 \ \ \ \quad \qquad if\ |x|<\delta, \\
\delta|x|-\frac{1}{2}\delta^2 \quad otherwise.\\
\end{array}\right.
\end{equation}

\subsection{Parameter Estimation Network}
\label{sec:pen}

Given the learned 3D region proposals, we can extrude the object points of interest from the whole point cloud. We do data augmentation to the cropped point sets by random flipping, random rotation and bounding box perturbation. Our PE-Net extends the PointNet~\cite{qi2017pointnet} structure to learn point-wise features and estimate the parameters of the amodal oriented 3D bounding box, including the box center coordinate $(x, y, z)$, the box size $(h, w, l)$ and the heading angle $\gamma$.

\noindent\textbf{Network Architecture.}
Inspired by PointNet~\cite{qi2017pointnet}, our PE-Net firstly employs a shared MLP on each point independently to extract local features, and subsequently a max-pooling layer is used to descend the dimension of features. Finally, several fully connected convolutional layers are adopted to get the output. T-Net, involving only a MLP, is employed for the centroid alignment of the object points. The architecture of our PE-Net is shown in Fig.~\ref{fig:pipeline}$(c)$.

\noindent\textbf{Stage-2 Loss Function.} 
To estimate the box center coordinates, we first normalize the extruded object points by subtracting the mean coordinate of points and then predict the residual box center coordinate, instead of directly predicting the original box center coordinate.
For the box size and orientation estimation, we use a hybrid formulation of classification and regression. We equally split $[0, 2\pi)$ into $N_H$ angle bins, and thus PE-Net will output $N_H$ scores for orientation classification and $N_H$ residual angles for regression. Based on the above analysis, our stage-2 loss $L_{stage-2}$ involves three components: center coordinate estimation, size estimation and orientation estimation, defined as
\begin{equation}
\begin{aligned}
L_{stage-2} =&L_{c1-reg} + L_{c2-reg} + L_{s-cls} +\\
&L_{s-reg} + L_{h-cls} + L_{h-reg} + L_{corner},
\end{aligned}
\label{eq:loss2}
\end{equation}
where the corner loss $L_{corner}$ is a function joining the center, size and orientation, proposed by~\cite{qi2018frustum}. 

As illustrated in Fig.~\ref{fig:pipeline}, at test time, a point cloud of size $N\times 4$ is firstly projected to a $128\times 512 \times 3$ front-view map which goes through PG-Net to generate 3D region proposals. And then the object points are extruded from the whole point cloud with 3D region proposals. Next, the object points go through PE-Net and then output the final bounding box parameters.


\begin{table*}
\renewcommand{\arraystretch}{1.0}
\caption{Performance comparison for 3D object detection on KITTI \textit{test} set with AP (\%) and inference time (ms).}
\label{table:comparison}
\centering
\begin{tabular}{|c|ccc|c|c|c|c|c|c|c|c|c|c|}
    \hline
    \multirow{2}{*}{Method} & \multirow{2}{*}{RGB} & \multirow{2}{*}{BEV} & \multirow{2}{*}{FV} &  \multirow{2}{*}{Time} &\multicolumn{3}{c|}{Car}  & \multicolumn{3}{c|}{Pedestrian} & \multicolumn{3}{c|}{Cyclist} \\  \cline{6-14}
     & & & & & Easy & Mod. & Hard & Easy & Mod. & Hard & Easy & Mod. & Hard \\ 
     \hline
    MonoPSR~\cite{ku2019monopsr} & \checkmark & & & 200 & 12.57 & 10.85 & 9.06 & 12.65 & 10.66 & 10.08 & 13.43 & 11.01 & 9.93\\
    Stereo R-CNN~\cite{li2019stereo} & \checkmark & & & 400 & 49.23 & 34.05 & 28.39 & - & - & - & - & - & - \\
     MV3D (LiDAR)~\cite{chen2017multi} & & \checkmark & \checkmark &240 & \textbf{66.77} & 52.73 & 51.31 & - & - & - & - & - & - \\
     BirdNet~\cite{beltran2018birdnet} & & \checkmark & & 110 & 14.75 & 13.44 & 12.04 & 14.31 & 11.80 & 10.55 & 18.35 & 12.43 & 11.88 \\ 
     RT3D~\cite{zeng2018rt3d} & & \checkmark & & 90 & 23.49 & 21.27 & 19.81 & - & - & - & - & - & - \\
     VeloFCN~\cite{li2016vehicle} & & & \checkmark & 1000 & 15.20 & 13.66 & 15.98 & - & - & - & - & - & - \\ 
     LMNet~\cite{minemura2018lmnet} & & & \checkmark & 20 & - & 15.24 & - & - & 11.46 & - & - & 3.23 & - \\
     \textbf{FVNet} & & & \checkmark & \textbf{12} & 65.43 & \textbf{57.34} & \textbf{51.85} & \textbf{42.01} & \textbf{34.02} & \textbf{28.43} & \textbf{38.03} & \textbf{24.58} & \textbf{22.10} \\ 
     \hline
\end{tabular}
\end{table*}

\section{Experiments}

In this section, we first introduce the KITTI dataset~\cite{geiger2013vision} and describe our experiment settings. Then our method is compared with some state-of-the-art 3D object detection methods on the KITTI benchmark, in terms of both accuracy and inference time. Finally, we show some representative visual results and analyze our framework.  


\subsection{Dataset and Settings}

\noindent\textbf{Dataset.}
We use the challenging KITTI object detection benchmark \cite{geiger2013vision} for both training and evaluation. KITTI provides camera images and raw point clouds, including $7,481$ sets for training and $7,518$ sets for testing. \textit{We only take the point clouds as input. The camera images are used for 3D bounding boxes visualization only.} Our method can handle the point clouds around $360^{\circ}$, but we only focus on the points within the camera's FOV for simplicity. We empirically set the maximum radial distance $R$ to $80$ meters.

The ``Pedestrian'' category and the ``Cyclist'' category are quite similar on the front-view maps because both categories contain human body and have small pixel sizes. Given this, we only detect two classes, the ``Car'' category and the ``Person'' category (the sum of the ``Pedestrian'' category and the ``Cyclist'' category) in the proposal generation stage, and then distinguish the ``Pedestrian'' category from the ``Cyclist'' category in the parameter estimation stage. Thanks to the finer features extracted from the object points, pedestrians and cyclists can be easily identified. We start from scratch to train the model weights over the preprocessed feature maps. 

\noindent\textbf{Implementation Details.}
We train our PG-Net with a mini-batch size of $32$ for $10^5$ iterations and we train our PE-Net with a mini-batch size of $64$ for $2\times 10^5$ iterations. It takes about $20$ hours for training PG-Net and $13$ hours for training PE-Net, with an NVIDIA 1080 Ti GPU and an i5-8600K 3.60GHz CPU. For data augmentation, we randomly crop and flip the front-view maps during PG-Net training, and apply random flip along $X$ axis and random rotation in $[-\pi/10, \pi/10]$ along the $Z$ axis during PE-Net training. We use non-maximal suppression (NMS) on front-view proposals with a threshold of $0.45$ to remove overlapped proposals.

\noindent\textbf{Evaluation Metric.}
Similar to previous works~\cite{li2016vehicle,chen2017multi, yang2018pixor}, we take the Average Precision (AP)~\cite{everingham2010pascal} as a standard for evaluation. Following the KITTI setting, we evaluate the performance of our proposed framework on three different cases: \textit{easy}, \textit{moderate} and \textit{hard}. The ``Car'' category is evaluated at $0.7$ Intersection-Over-Union (IoU) while the ``Pedestrian'' and ``Cyclist'' categories are at $0.5$ IoU. Similar to~\cite{li2016vehicle, yang2018pixor}, we ignore other categories in KITTI such as ``Van'' and ``Person\_sitting''.


\subsection{Comparison with State-of-the-Art Methods}

We compare our approach with state-of-the-art methods \cite{ku2019monopsr,li2019stereo,chen2017multi,beltran2018birdnet,zeng2018rt3d,li2016vehicle,minemura2018lmnet}, which are divided into two groups depending on the input (i.e., point clouds or camera images). One group consists of MonoPSR~\cite{ku2019monopsr} (Mono-based) and Stereo R-CNN~\cite{li2019stereo} (Stereo-based) which process camera images with RGB information. 
The other group includes MV3D (LiDAR)~\cite{chen2017multi}, BirdNet~\cite{beltran2018birdnet}, RT3D~\cite{zeng2018rt3d}, VeloFCN~\cite{li2016vehicle} and LMNet~\cite{minemura2018lmnet} which are based on point clouds only. All previous results are borrowed from the KITTI leadbord. 

\begin{figure*}
\begin{center}
	\includegraphics[width=1\linewidth]{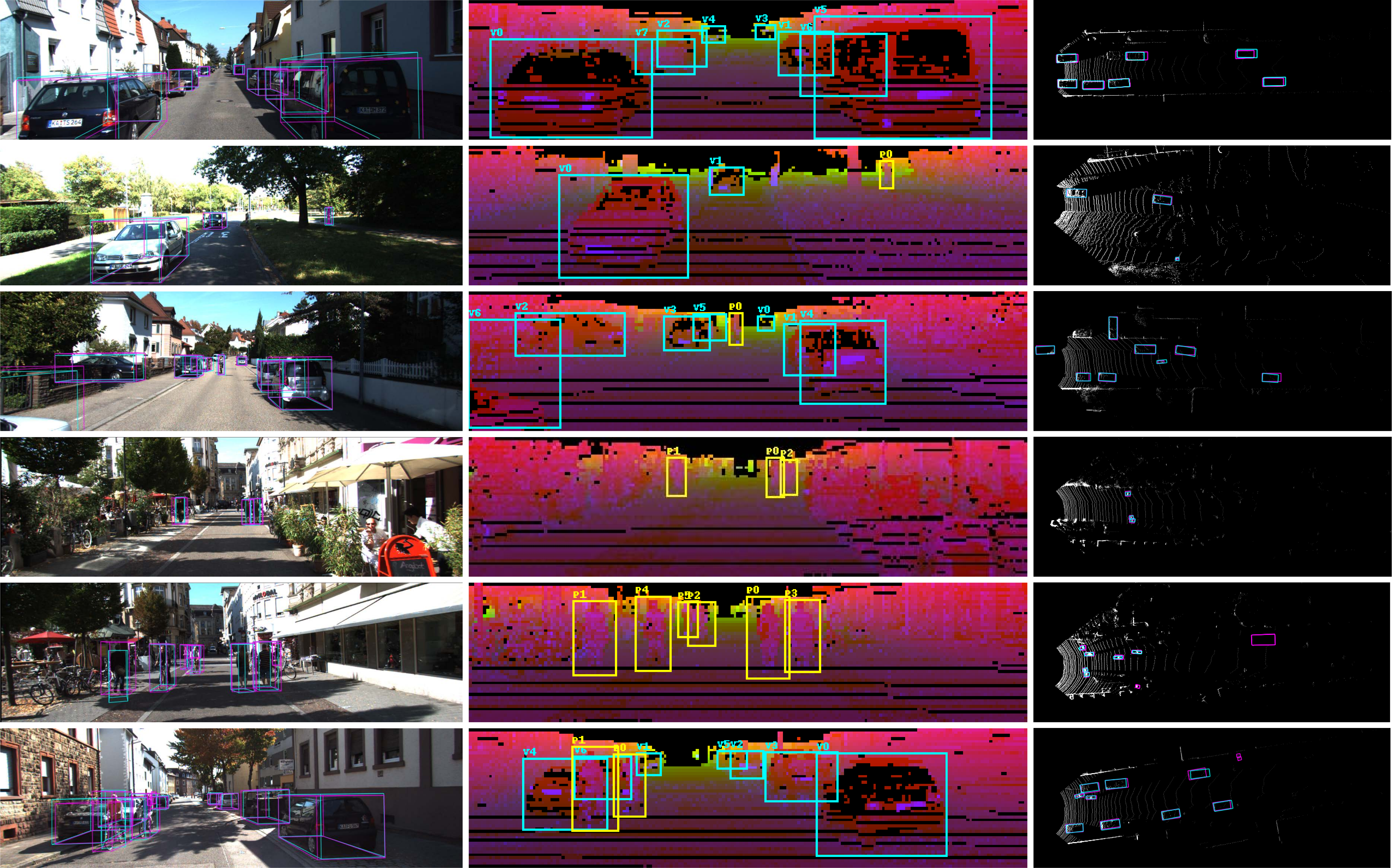}
\end{center}
   \caption{Visual results of 3D object detection by our FVNet. The left column shows the camera images with detected bounding box in blue and ground truth in red. The middle column shows the front-view maps with car predictions in blue and person predictions in yellow, and the right column shows the bird's-eye-view maps with consistent colors with the left column (camera images). The notation attached to each bounding box denotes the detected object category, in which ``V'' and ``P'' indicate cars and people (including pedestrians and cyclists), respectively.}
\label{fig:vis}
\end{figure*}

Table~\ref{table:comparison} reports the AP results and inference time of these methods. Compared with the camera-images-based methods, our method FVNet achieves significant better results despite using the raw point clouds only. Overall, our method performs best, except the car detection in \textit{easy} setting by MV3D (LiDAR) which employs both front-view and bird's-eye-view maps. Notice that most of the existing methods report no or very limited detection results on Pedestrian or Cyclist categories due to the challenge of these two categories. In contrast, our approach obtains competitive results, especially for the easy setting of Pedestrian ($42.01\%$ compared to $14.31\%$ of BirdNet \cite{beltran2018birdnet}). This demonstrates that our framework is effective for detecting small objects like pedestrians or cyclists.

It can be seen that our proposed method achieves the highest inference speed with only $12$ms per sample. This is because that our framework generates high-quality 3D proposals from front-view maps directly, without the need of the instance segmentation network. Although LMNet \cite{minemura2018lmnet} shows a comparable inference speed, its accuracy is remarkably lower than ours. MV3D (LiDAR) \cite{chen2017multi} performs generally the second best in accuracy but costs $20$ times of inference time to ours.

\begin{figure*}
\centering
	\includegraphics[width=1\linewidth]{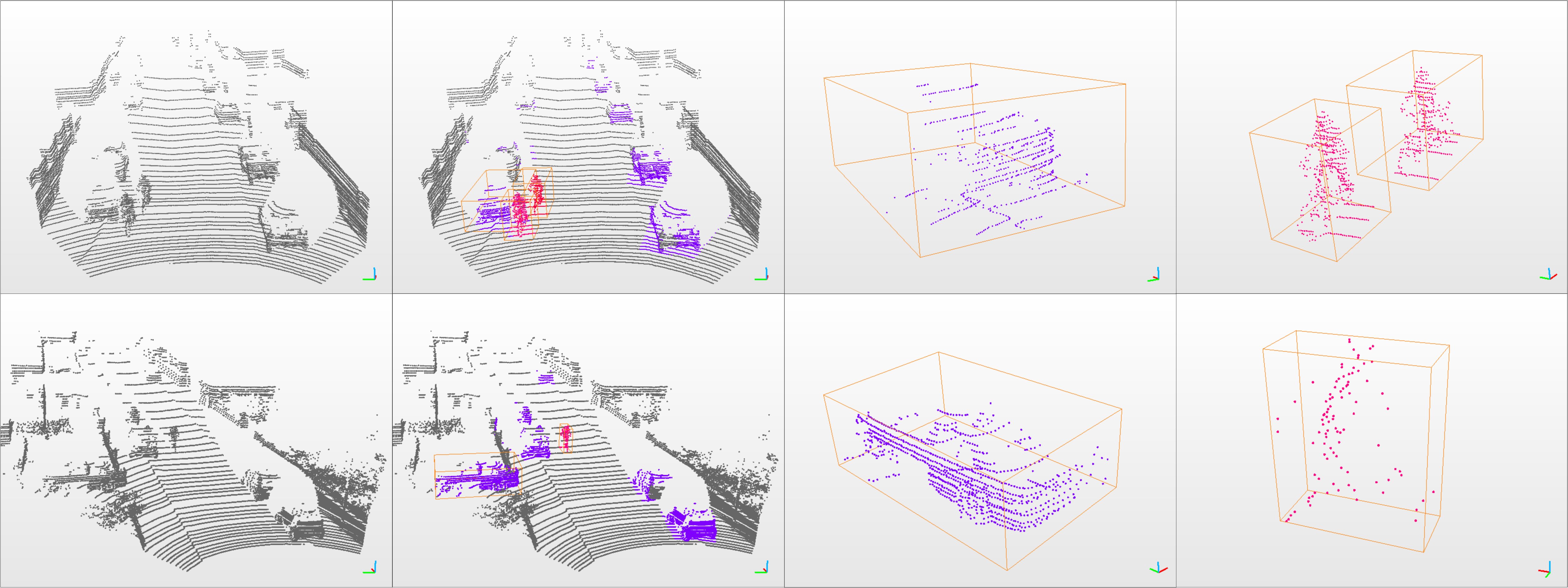}
   \caption{Visualization of 3D proposals generated by our PG-Net. The first column shows the raw point clouds with noise and outliers. The second column displays the 3D proposals in raw point clouds. The third column visualizes the predicted 3D proposals of cars, and the last column demonstrates the predicted 3D proposals of persons. The bounding boxes are to highlight the 3D proposals. }
\label{fig:proposals}
\end{figure*}

\subsection{Visual Results}
We present six representative 3D object detection examples in Fig.~\ref{fig:vis}.
It can be seen that our framework FVNet provides accurate 3D bounding boxes in all object categories. The top row shows that our method can easily deal with the dense cars. It can even detect small cars at a  distance, such as the instances ``V3'' and ``V4'' in the corresponding front-view map. The second row demonstrates that our network can take small objects in the dark into account, for example, the instance ``P0'', a pedestrian under the shade of trees which is almost invisible in the bird's-eye-view map. The third row indicates that even the object is hidden from view, our network can still predict its accurate size, such as the ``V6'' instance in the corresponding front-view map. The fourth and fifth rows show that our network has the ability to distinguish people from the crowd, including pedestrians and cyclists. From the last row we can see that when people overlap cars with a short distance, our network can separate them from each other clearly. 
In general, the proposed FVNet is applicable and robust in various cases, such as small objects, poor light conditions and slight occlusions.

\subsection{Model Analysis}
We analyze our framework from the aspects of 3D proposals, multi-scale detector and failure case. 

\begin{table}
\renewcommand{\arraystretch}{1.0}
\caption{Evaluation on the Car category of KITTI \textit{test} set using single-scale or multi-scale detectors.}
\label{table:scale}
\centering
\begin{tabular}{|*{6}{c|}}
\hline
Setting&Scale &2D&AOS&BEV&3D\\
\hline
\multirow{2}*{Easy} 
&Single& 75.42 & 74.22 & 63.79 & 50.97  \\
&Multi& \textbf{86.14} & \textbf{85.94} & \textbf{78.04} & \textbf{65.43}  \\
\hline
\multirow{2}*{Mod.} 
&Single& 69.72 & 67.96 & 60.59 & 47.64  \\
&Multi& \textbf{77.19} & \textbf{76.84} & \textbf{65.03} & \textbf{57.34}  \\
\hline
\multirow{2}*{Hard} 
&Single& 63.66 & 61.78 & 56.19 & 44.56  \\
&Multi& \textbf{69.27} & \textbf{68.90} & \textbf{57.89} & \textbf{51.85} \\
\hline
\end{tabular}
\end{table}

\noindent\textbf{3D Proposals.}
Fig.~\ref{fig:proposals} visualizes two examples of 3D proposals. Notice that these proposals are directly generated by our PG-Net rather than the final 3D bounding box predictions.  
It can be observed that the 3D proposals predicted from raw point clouds are convincing and reasonable. In the top row, persons with few pavements are included in the 3D proposals. In the last row, we can observe that the cars are correctly detected with clear shape of wheels. Since the results are comparable to those by instance segmentation, it is sufficient to accurately generate the final results with our 3D proposals.

\noindent\textbf{Multi-Scale Detector.}
To prove the effectiveness to detect small objects at a distance of the proposed multi-scale detector, we also implement a single-scale variant of our framework. We compare our multi-scale detector with the single-scale detector on the Car category, shown in Table \ref{table:scale}. It can be seen that the multi-scale detector achieves noticeably better accuracy, with margins of $14.5\%$, $9.7\%$ and $7.3\%$ in the settings of easy, moderate and hard for 3D object detection, respectively. This reveals that the proposed multi-scale detector is more effective than the sing-scale version in detecting small objects.

\begin{figure}
\centering
	\includegraphics[width=1\linewidth]{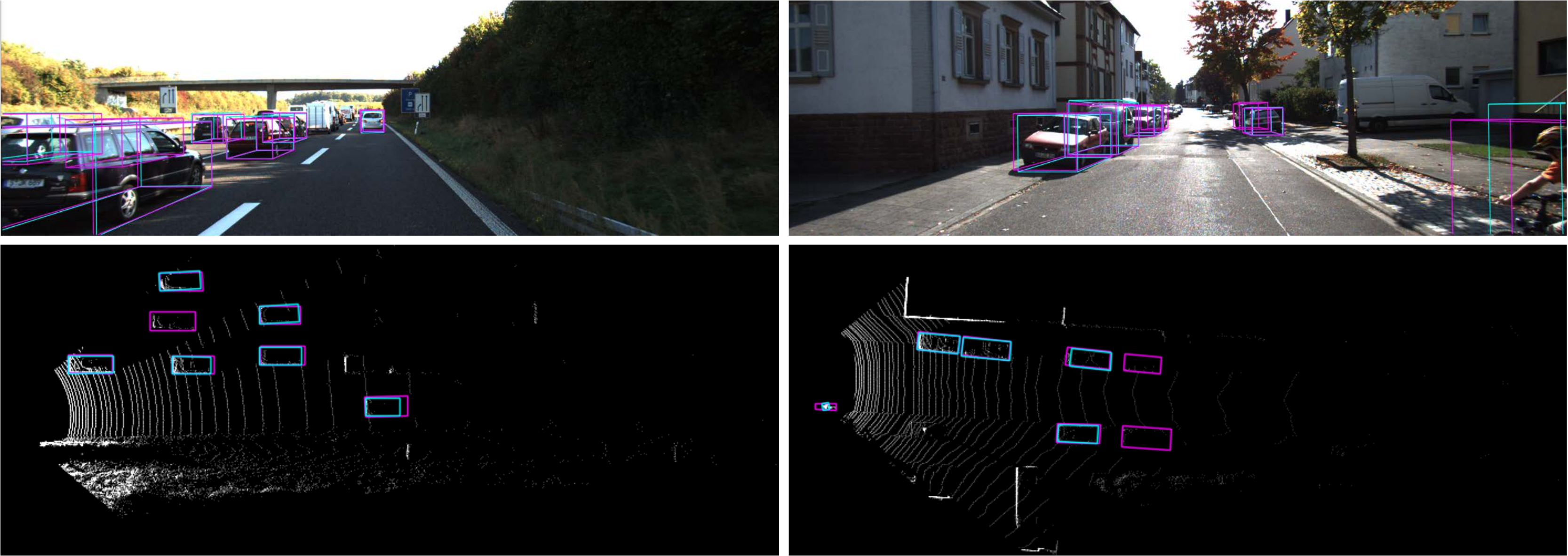}
   \caption{Failure cases: some objects are severely occluded. The first and second rows show results for camera images and the corresponding bird’s-eye-view maps, respectively. The detected bounding boxes and the corresponding ground-truth boxes are colored in blue and red, respectively.  }
\label{fig:failure}
\end{figure}

\noindent\textbf{Failure Case.}
Despite the proven accuracy and robustness, our method may fail in the case of severe occlusions. Some examples are showed in Fig.~\ref{fig:failure}, where some objects are occluded and the points are extremely sparse. As a result, the generated front-view maps are not reliable. This issue can be potentially alleviated by using the bird's-eye-view maps, which we would like to investigate in the future. 

\section{Conclusion}
In this paper, we have introduced a novel learning based approach for 3D object detection, which comprises of the proposal generation network and the parameter estimation network. Our method takes only point clouds as input, without the need of fusing other information like camera images. The results of extensive experiments show that our method outperforms state-of-the-art image-based and projection-based methods with only point clouds as input, in terms of accuracy and inference time. Above all, our approach achieves real-time performance with 12ms per point cloud sample, and is significantly faster than other techniques. 



\section*{Acknowledgment}
Thanks to the National Natural Science Foundation of China (No. 61972157), Science and Technology Commission of Shanghai Municipality Program (No. 18D1205903) and the National Social Science Foundation of China (No. 18ZD22) for funding.

\bibliography{references} 
\bibliographystyle{IEEEtran}

\end{document}